\documentclass[10pt]{article}

\usepackage{amsmath,amssymb,amsthm}
\usepackage{url}
\usepackage{graphicx}
\usepackage{color}
\usepackage{times}

\usepackage{float}
\usepackage{subfig}

\usepackage{alltt}
\usepackage[boxed]{algorithm}
\usepackage{algorithmic}
\usepackage{array}
\usepackage{comment}
\usepackage{multirow}
\usepackage{paralist}

\renewenvironment{thebibliography}[1]{%
  \begin{oldthebibliography}{#1}%
    \setlength{\parskip}{0ex}%
    \setlength{\itemsep}{0ex}%
}%
{%
  \end{oldthebibliography}%
}

\makeatletter
\def\url@smallurlstyle{%
  \@ifundefined{selectfont}{\def\UrlFont{\sf}}{\def\UrlFont{\footnotesize\ttfamily}}}
\makeatother
\usepackage[left=1.25in, top=1in, bottom=1in, right=1.25in]{geometry}

\allowdisplaybreaks

\newcolumntype{C}[1]{>{\centering\arraybackslash}p{#1}}

\newcommand{\myalign}[1]{\begin{align*}#1\end{align*}}
\newcommand{\mytt}[1]{{\small\begin{alltt}#1\end{alltt}}}

\newcommand{\blank}[1]{}
\newcommand{\typewriter}[1]{{\small\texttt{#1}}}

\title{Dyadic Prediction Using a Latent Feature Log-Linear Model}
\author{Aditya Krishna Menon \\ Department of Computer Science and Engineering \\ University of California, San Diego \\ \texttt{akmenon@cs.ucsd.edu} 
\and Charles Elkan \\ Department of Computer Science and Engineering \\ University of California, San Diego \\ \texttt{elkan@cs.ucsd.edu}}
\date\today

\begin{document}

\maketitle

\begin{abstract}
In dyadic prediction, labels must be predicted for pairs (dyads) whose members possess unique identifiers and, sometimes, additional features called side-information. Special cases of this problem include collaborative filtering and link prediction. We present the first model for dyadic prediction that satisfies several important desiderata:
(i) labels may be ordinal or nominal,
(ii) side-information can be easily exploited if present,
(iii) with or without side-information, latent features are inferred for dyad members,
(iv) it is resistant to sample-selection bias,
(v) it can learn well-calibrated probabilities, and
(vi) it can scale to very large datasets.
To our knowledge, no existing method satisfies all the above criteria. In particular, many methods assume that the labels are ordinal and ignore side-information when it is present. Experimental results show that the new method is competitive with state-of-the-art methods for the special cases of collaborative filtering and link prediction, and that it makes accurate predictions on nominal data.
\end{abstract}

\section{ {Limitations of existing dyadic prediction methods}}

In dyadic prediction, the training set consists of {pairs} of objects $\{(r_i, c_i)\}_{i=1}^n$, called \emph{dyads}, with associated labels $\{ y_i \}_{i=1}^n$. 
The task is to predict labels for unobserved dyads i.e.\ for pairs $(r', c')$ that do not appear in the training set. A well-studied special case of dyadic prediction is where the dyads refer to (user, item) pairs, and the labels are a user's numeric rating of an item, e.g.\ on a scale of $1$ to $5$. The task of predicting the ratings of unobserved (user, item) pairs is an example of \emph{collaborative filtering}, where the goal is to recommend to users new items they might like, e.g.\ items we predict the user will rate $\geq 4$ stars.

Existing methods for dyadic prediction are limited in one or more ways, which we now discuss. The most common limitation  is that they assume\footnote{This assumption is reflected in the loss function used for training, typically mean absolute or mean square error, which further imposes a numeric structure on the labels.} the observed labels have an \emph{ordinal} scale: a rating of $1$ star is less than a rating of $2$ stars, for example. 
But in many real-world applications of dyadic prediction, the outcomes are \emph{nominal}, or unordered; for example, suppose an online store has information about customers' interactions with their products, with possible outcomes being \{viewed, purchased, returned\}. On such datasets, imposing an ordinal structure is inappropriate and leads to poor accuracy. An ideal dyadic prediction model should be flexible enough to handle both types of label.

Another issue is that many models only exploit dyad members' unique identifiers or their additional attributes (called \emph{side-information} or covariates). 
A model that uses only one of these distinct pieces of information has limited applicability. One that requires side-information is obviously inapplicable when only unique identifiers are present. Similarly, one that only uses members' unique identifiers is limited in the \emph{cold-start} setting, where the test set contains a dyad $(r',c')$, where at least one of $r'$ or $c'$ is not present in any training set dyad. An example of this setting is predicting how a user will rate a movie that is yet to be released, and thus has no existing ratings.


A characteristic of dyadic prediction is \emph{sample-selection bias}, meaning that the training and test set follow different distributions. In collaborative filtering for example, it has been observed that users are more likely to provide ratings for items that they like, which biases the training set. This bias poses a challenge for generative probabilistic models, but is handled by their discriminative counterparts \cite{GenerativeSSBias} and also non-probabilistic models. The downside with the latter is that learning probabilities, and more specifically \emph{well-calibrated} probabilities \cite{Calibrated}, is useful because it lets us use the model as part of a larger framework; for example, we might send a customer a free sample of a product if the cost of sending it is less than the expected revenue it induces (due to it influencing her future purchases).

Finally, it is essential for a dyadic prediction model to scale to very large datasets, as these are common in practical applications. If training cannot be done efficiently, the applicability of a model is severely limited.

In this paper, we propose the first model that addresses all the above issues. Specifically, its salient features are: (i) it can handle both nominal and ordinal labels, (ii) side-information can be incorporated seamlessly if it present, (iii) the model infers latent features for dyad elements even in the absence of side-information, (iv) it is a discriminative method, and so is resistant to sample-selection bias, (v) it can produce well-calibrated probabilities, and (vi) it can scale to very large datasets. Another appealing property of our model is its conceptual and technical simplicity: we use a log-linear model where the log-odds are approximated by a low-rank matrix. Based on this, we call our model the LFL (for \emph{latent feature log-linear}). The flexibility of a log-linear model allows us to add side-information easily, and lets us handle both nominal and ordinal labels by just changing the loss function for training. By virtue of it being a discriminative probabilistic method, we automatically obtain resistance to sample-selection bias and can output well-calibrated probabilities. Experimental results show that the model does as well as existing methods for collaborative filtering and link prediction, which demonstrates that we do not lose predictive power by addressing a more general problem. We also show that we can learn from nominal data, and that by capturing side-information we can give good predictions in the cold-start setting.

\section{ {Background}}

In this section, we define dyadic prediction in terms of matrix completion. We define the desiderata for a dyadic prediction model, and why they are not all met by existing methods.

\subsection{ {The dyadic prediction problem}}

In dyadic prediction, our training set is $\{((r_i,c_i),y_i)\}_{i=1}^n$, where the pairs $(r_i,c_i)$ are known as \emph{dyads}. Usually, but not necessarily, the only information we have about each element in the pair is a unique identifier. Our goal is to predict the label $y$ for an unobserved dyad $(r',c')$. We can interpret this task as a form of \emph{matrix completion} by associating the training set with a matrix $M \in \mathcal{Y}^{|\mathcal{R}| \times |\mathcal{C}|}$, where $r_i \in \mathcal{R},c_i \in \mathcal{C}, y_i \in \mathcal{Y}$. Each row of $M$ is associated with some $r \in \mathcal{R}$ and each column with some $c \in \mathcal{C}$, so the training data is simply a subset of observed entries in $M$. Our task is to fill in the missing entries of $M$. Based on this interpretation, we will henceforth refer to $r_i$ as a \emph{row object} and $c_j$ as a \emph{column object}.

One approach to solving the matrix completion problem is to treat it as $m$ separate learning tasks, one for each row object $r_i$. When there are no features for the row and column objects, we can use the entries of other rows as features. We are assuming here that the entries for other row objects contain predictive information. 
Techniques along these lines have been proposed in \cite{PRank,MarginRank}. There are a few problems with this approach: (i) treating each learning task separately is time-consuming, (ii) we have to come up with a heuristic to fill in missing entries when constructing features, and (iii) by making each learning task completely separate, we are possibly losing a lot of shared information among the row objects. A recourse for problem (iii) might be to view the problem in a multi-task setting - indeed, one can interpret some dyadic prediction methods from a multi-task viewpoint \cite{MMMF-MT} - but the other issues remain.

Given the difficulties with artificially constructing features, a natural recourse is to try to discover \emph{latent features}. Ignoring missing entries for the moment, and supposing that $\mathcal{Y} = \mathbb{R}$, the classical way to learn such latent features from a matrix $M$ is through the singular value decomposition (SVD). This expresses $M$ as the product of matrices $U$ and $V$, which are representations of row and column objects in some latent feature space. In the case of collaborative filtering, this has a natural interpretation as being users' and items' propensities towards certain features (e.g.\ whether the item is a status symbol, whether the user likes foreign films). Of course, in matrix completion we do not know the entire matrix $M$, which means the SVD cannot be applied as-is. In this setting we can restate the task as finding a low-rank approximation that agrees with $M$ only on its observed entries; the hope is that if there are enough observed entries, this approximation will generalize to the missing entries as well. This problem is non-convex - there are many local optima corresponding to matrices that are not merely isomorphic upto rotation \cite{WeightedLR} - and so it is difficult to solve. Methods like MMMF \cite{MMMF}, discussed later, have been proposed to make this learning tractable.

\subsection{Desiderata for a dyadic prediction model}
\label{sec:dyadic-desiderata}

As mentioned in the introduction, there are several desiderata for a dyadic prediction model:

    \textbf{Agnostic to nature of labels}. Labels in a dyadic prediction task may be ordinal, for example star ratings for a (user, movie) pair, or nominal, for example one of \{bought, viewed, returned\} for a (user, product) pair. Ideally, a model should handle both types of labels.
    
    \textbf{Exploiting side-information}. It is important for a model to be able to use side-information when it is provided; otherwise, it is severely limited in the aforementioned cold-start setting.
    
    \textbf{Learning latent features}. It is desirable for a model to infer latent features for dyad elements just from the identities of row/column objects; otherwise, we are severely limited when we have no side-information. Aside from endowing the model with stronger predictive power, it also helps us understand the data better. (See the experimental results in \S\ref{sec:nips-coauthor}, for example, where we use the latent features to cluster the data.)
    
    \textbf{Resists sample-selection bias}. {Sample-selection bias} is the setting where the training and test sets follow different distributions. This is manifested in movie rating prediction for example, where users are generally more likely to provide ratings for movies that they like. This bias poses a challenge for generative models, because they need to model the joint distribution of examples and labels. A discriminative model, in contrast, only focusses on conditional probabilities of labels, and so is inherently robust against this bias \cite{GenerativeSSBias}.    
    
    \textbf{Well-calibrated probabilities}. Often, a classifier is only a sub-model in a larger framework. In this setting, having the classifier output well-calibrated probabilities \cite{Calibrated} rather than just ordinal scores helps us make decision-theoretically optimal choices. For example, we might decide to send a customer a free sample of a product if we determine the cost of sending it is less than the expected revenue the action will bring about, due to it encouraging her to purchase similar products.

    \textbf{Fast training}. Large datasets are common for practical applications of dyadic prediction such as collaborative filtering, so scalable training algorithms are essential.

To the best of our knowledge, the log-linear model proposed in this paper, which we call the LFL, is the first method to meet all these criteria. We now briefly discuss existing methods for dyadic prediction, pointing out which issues they fail to address.

\subsection{ {Existing dyadic prediction methods}}

We will focus here on models for two well-studied special cases of dyadic prediction, namely collaborative filtering and link prediction. We wish to note at the outset that the focus of this paper is on models for the general dyadic prediction; as such, we are not interested in tuning our model to the specifics of a collaborative filtering problem, which is the focus of many published results in the literature. We note that it is conceptually simple to incorporate many of the tricks for collaborative filtering into our LFL model. (This is discussed further in \S\ref{sec:bias-weight}.)

\textbf{Matrix factorization}. The idea of learning \emph{latent features} through matrix factorization has been very successful in collaborative filtering. A popular example of this is maximum margin matrix factorization (MMMF) \cite{MMMF}, where the idea is to approximate the input matrix $M$ (containing missing entries) by a matrix $X$ whose complexity is controlled by a convex approximation to the rank:
$$\min_X \sum_{i,j \in I} \ell(X_{ij}, M_{ij}) + \lambda ||X||_{\Sigma},$$
where $I$ is the set of observed entries in $M$, $\ell$ is a modified hinge-loss, and $||\cdot||_\Sigma$ is the trace norm of a matrix (the sum of its singular values, also known as the nuclear norm). Penalizing the trace norm favours matrices $X$ that are explained by a few latent factors i.e.\ if $X = UV^T$ for rank $k$ matrices $U,V$, then $||U||_F^2 + ||V||_F^2$ is small. The above formulation requires solving an SDP and so is not very scalable. A faster alternative (which sacrifices convexity) is to directly learn $U, V$ with a gradient-based method \cite{FastMMMF}. Probabilistic extensions of MMMF such as \cite{PMF,BPMF} give a Bayesian treatment of the problem, and obtain higher accuracy. These were further extended in \cite{GPMF}, which interprets the problem in a Gaussian process framework to learn a \emph{nonlinear} matrix factorization. A related result is \cite{NPCA}, where the focus is on a nonparametric model with the number of latent factors $k$ determined automatically.

As noted earlier, a significant amount of research has focussed on improving the performance of matrix factorization methods for the collaborative filtering setting. One such line of work has studied how to combine \emph{neighbourhood models} with standard matrix factorization. A neighbourhood model is based on the following simple idea: similar users tend to rate movies similarly. These models are good at capturing local effects in collaborative filtering data, and have been shown to improve performance of standard matrix factorization models. Recent models based on this hybrid approach are \cite{Takacs} and \cite{Koren}. The latter also gives a way to incorporate \emph{implicit feedback} in collaborative filtering, where we exploit information about \emph{which} movies a user rated, even if we do not know the actual rating: the mere fact that a user has rated a movie usually increases the chance the (s)he likes it. However, it does not address the issue of incorporating more general forms of side-information. We note that it is conceptually simple to add in neighbourhood information to our model using techniques similar to these papers; we emphasize once more that our goal is to address a broader set of concerns than just improving accuracy of collaborative filtering methods.

All matrix factorization methods assume that the input labels are real, and thus ordinal. Almost all of them do not incorporate side-information. Recent exceptions are \cite{GPMF, OperatorEstimation}; to our knowledge the latter has not been tested extensively on a number of datasets.

\textbf{Boltzmann machines}. Restricted Boltzmann machines (RBMs) have enjoyed some success for collaborative filtering \cite{RBM-CF}. The probability model for an RBM is a Gibbs distribution, where, we model
$$p(x,y,h;w) = \frac{\exp \Psi(x,y,h)}{\sum_{x',y',h'} \exp \Psi(x,y,h)}.$$
Here, $x,y$ are the inputs and labels, $h$ a number of binary-valued hidden units, and $\Psi$ is some linear function of its inputs.

An important difference between the RBM and a log-linear model is that the RBM is a \emph{generative model}. This makes training more difficult, because we are solving a more complex modelling task. Perhaps more importantly, as we noted earlier, generative training makes a model susceptible to sample-selection bias. A discriminative form of the RBM was proposed in \cite{DiscRBM}, but it has not been applied in a collaborative filtering context. Further, to the best of our knowledge, models based on the RBM have not been extended to incorporate side-information, and have not been applied to datasets with nominal labels. Finally, training these models involves approximation schemes such as contrastive divergence, as the necessary gradients are difficult to compute. Our model does not suffer from these issues.

\textbf{Link prediction models}. In link prediction, the input is the adjacency matrix $M$ of a graph with some missing entries, which we want to fill in. This is a dyadic prediction problem where both objects in the dyad belong to the same space, e.g.\ people in a social network. Additional (optional) constraints are that the graph is undirected and unweighted i.e.\ $M$ is binary and symmetric. Two link prediction models that are relevant for our work are \cite{HiddenLR} and \cite{IBP}. \cite{HiddenLR} handles the case of binary $M$ using logistic regression, where the log-odds are modelled by a low-rank matrix approximation. This is similar to the model we propose, except that our training procedure is considerably simpler than the MCMC scheme used in the paper, and also our model addresses the general dyadic prediction task, with binary link prediction as a special case. \cite{IBP} also uses logistic regression, but with two important distinctions from \cite{HiddenLR}: (i) the matrix decomposition involves binary matrices, indicating the presence of a particular latent feature, and (ii) the decomposition is nonparametric, so the number of latent factors need be specified a-priori. Training in this model is involved, and so it is not clear that it scales to very large datasets.

\textbf{Other models}. Early work for dyadic prediction took a mixture modelling approach. In the aspect model \cite{Aspect}, we assume that there are some underlying groups or clusters for the dyads $(r,c)$, and so each example in our training set is a draw from a set of these clusters. Learning can be done using expectation maximization to account for the latent groups for each dyad. In a collaborative filtering context, these models have only had limited success, in part because they are quite restrictive in terms of modelling the generation of each dyad \cite{Marlin}.

A popular approach to dyadic prediction is to \emph{co-cluster} the input matrix $M$, meaning we simultaneously cluster its rows and columns (see e.g.\ \cite{PDLF}). 
These models are reliant on side-information for each of the dyad elements, however.

\cite{InfiniteBinary} proposes a hierarchical Bayesian model that is similar to \cite{IBP}, but does not focus exclusively on link prediction. In this sense, it has similar goals to our model. On the technical side, as the treatment is Bayesian, training and inference require approximation techniques. Also, the model does not exploit side-information when it is present. Finally, our viewpoint emphasises the fact that our model can easily optimize multiple objective functions for ordinal tasks, meaning it can be tuned for the specifics of the problem at hand.

\textbf{Summary}. We close this section with Table \ref{tbl:methods-summary}, which summarizes various existing methods in the literature in terms of whether they possess each of the six desiderata we listed in \S\ref{sec:dyadic-desiderata}. We see that the LFL model proposed in this paper is the first method that meets \emph{all} six desiderata.

\begin{table}[htb]
	\centering
	\begin{tabular}{|cc|ccp{1.5cm}p{1.25cm}p{1.25cm}p{1.25cm}|}
	\hline
	\textbf{Method} & \textbf{Reference} & Nominal? & Side-info? & Latent features? & Resists SS bias? & Calibrated probs.? & Fast training? \\
	\hline
	MMMF & \cite{MMMF} & No & No & Yes & Yes & No & No \\
	PMF & \cite{PMF} & No & No & Yes & No & No & No \\	
	NPCA & \cite{NPCA} & No & No & Yes & No & No & No \\	
	GPMF & \cite{GPMF} & No & Yes & Yes & No & No & Yes \\
	BRISMF & \cite{Takacs} & No & No & Yes & Yes & No & Yes \\
	FactNgbr & \cite{Koren} & No & No & Yes & Yes & No & Yes \\			
	RBM & \cite{RBM} & No & No & Yes & No & No & No \\
	\hline
	LFL & This paper & Yes & Yes & Yes & Yes & Yes & Yes \\
	\hline
	\end{tabular}
	
	\caption{Summary of various methods in the literature in terms of meeting the six desiderata from \S\ref{sec:dyadic-desiderata}. The desiderata are presented in the same order as in \S\ref{sec:dyadic-desiderata}.}
	\label{tbl:methods-summary}
\end{table}

\section{ {The latent feature log-linear model}}

In this section, we describe a simple log-linear model for dyadic prediction, and explain why it is limited. We address this limitation by adding latent features, yielding our latent feature log-linear model (LFL). We then discuss how to make predictions with and train our latent feature model for the nominal and ordinal settings.

\subsection{ {A simple log-linear model}}

Given an observation $x \in \mathcal{X}$ and label $y \in \mathcal{Y}$, a log-linear model represents the conditional distribution $p(y|x)$ via
$$p(y|x;w) = \frac{\exp\left[ \sum_i w_i f_i(x,y) \right]}{\sum_{y'} \exp\left[ \sum_j w_j f_j(x,y') \right]}.$$
Here, $w$ is a vector of real-valued weights to be learned. The functions $f_i : \mathcal{X} \times \mathcal{Y} \to \mathbb{R}$ are called \emph{feature functions}, and measure interactions between the inputs and labels. This model can be viewed as an extension of multinomial logistic regression. 

Recall that our inputs in a dyadic prediction task are $(x, y)$, where $x = (r, c)$ is a dyad and $y$ a label. We will use $r(x)$ and $c(x)$ to denote the respective elements in $x$. Suppose there is no side-information; then, $r,c$ are just indices into sets $\mathcal{R}, \mathcal{C}$ denoting the space of row and column objects respectively, and $\mathcal{X} = \mathcal{R} \times \mathcal{C}$. We apply a log-linear model for $p(y|x;w)$ with three sets of feature functions:
\begin{compactitem}
    \item $\text{for all } r \in \mathcal{R}, y' \in \mathcal{Y}$, let $f^{(1)}_{ry'}(x,y) = \mathbf{1}[r(x) = r, y = y']$
    \item $\text{for all } c \in \mathcal{C}, y' \in \mathcal{Y}$, let $f^{(2)}_{cy'}(x,y) = \mathbf{1}[c(x) = c, y = y']$
    \item $\text{for all } y' \in \mathcal{Y}$, let $f^{(3)}_{y'}(x,y) = \mathbf{1}[y = y']$
\end{compactitem}
where $\mathbf{1}[\cdot]$ denotes an indicator function. We can think of the feature functions as inducing a weight for each object-label pair. Specifically, split the weight vector into three components: $\alpha \in \mathbb{R}^{|\mathcal{R}| \times |\mathcal{Y}|}$, $\beta \in \mathbb{R}^{|\mathcal{C}| \times |\mathcal{Y}|}$ and $\gamma \in \mathbb{R}^{|\mathcal{Y}|}$. Then, the log-linear model defined by the feature functions is
\begin{equation}
\label{eqn:loglinear}
p(y|x;w) \propto {\exp\left[ \alpha_{r(x)y} + \beta_{c(x)y} + \gamma_{y} \right]}
\end{equation}

This model is conceptually simple, but it is limited in expressiveness. The most serious problem with the model is that it only learns \emph{propensities} of row and column objects towards a particular outcome
, and does not capture information about interactions between the two objects. To see an example of this, fix a row object $r_1$ and consider dyads of the form $(r_1, c)$. From Equation \ref{eqn:loglinear}, for some fixed outcome $y$, the ranking of all $c$'s in decreasing order of $p(y|(r_1, c);w)$ depends \emph{only} on $\beta_{cy}$. That means we get the exact same ranking of $c$'s for a different row object $r_2$.

\subsection{ {A richer latent feature model}}

The limitation of the previous model was the lack of interaction between row and column objects. We would like to have weights that capture more complex interactions. To do this, consider the model
\begin{equation}
\label{eqn:latent-loglinear}
p(y|x;w) = \frac{\exp\left( \sum_{i=1}^k \alpha^y_{r(x)i} \beta^y_{c(x)i} \right)}{\sum_{y'} \exp\left( \sum_{i=1}^k \alpha^{y'}_{r(x)i} \beta^{y'}_{c(x)i} \right)}
\end{equation}
where $k$ is some constant, denoting the number of \emph{latent factors} we learn from the training data. (The bias $\gamma_y$ is dealt with in the next section.) For each outcome $y$, $\alpha^y_{ri}$ represents a weight for row object $r$ on some latent feature $i$, and similarly for $\beta^y_{ci}$. So, we are learning tensors $\alpha \in \mathbb{R}^{|\mathcal{Y}| \times |\mathcal{R}| \times k}$ and $\beta \in \mathbb{R}^{|\mathcal{Y}| \times |\mathcal{C}| \times k}$. It is easy to check this model is not subject to the ranking problem discussed in the previous section.

Before discussing the intuition for these weights, we take care of a technical point to make further exposition clearer. In multinomial logistic regression, it is standard to fix one category to be the ``base'', which defines a scale for the other categories' weights. This eliminates the need to learn the weights for one category, and improves the {identifiability} of the learned model. In our model, we can specify the final outcome (or indeed any arbitrary one) to be our base category. Suppose without loss of generality that $y \in \{ 1, 2, \ldots, R \}$. Then, for $y \neq R$, we are specifying
$$p(y | x; w) = \frac{\exp(\alpha^y_{r(x):} (\beta^y_{c(x):})^T)}{1 + \sum_{y \neq R}\exp(\alpha^y_{r(x):} (\beta^y_{c(x):})^T)}.$$
This is equivalent to Equation \ref{eqn:latent-loglinear} where we enforce $\alpha^R, \beta^R \equiv 0$. We used this scheme for training our model.

Now we see why we call $\alpha, \beta$ latent weights. To do this, let us focus on the case $|\mathcal{Y}| = 2$ i.e.\ a logistic regression model. Let the outcomes be $y = 1$ and $y = 0$ without loss of generality. Following the above discussion, let $y = 0$ be the base class. In a slight abuse of notation, let $\alpha, \beta$ denote the matrices $\alpha^1, \beta^1$. Then, in our model the log-odds are
$$\log \frac{p(y=1|x;w)}{p(y=0|x;w)} = \alpha_{r(x):} (\beta_{c(x):})^T.$$
Defining a matrix $P$ whose entries are ${P}_{rc} = p(y=1|(r,c);w)$, we have the matrix decomposition
$P = \sigma(\alpha \beta^T)$, 
where $\sigma(\cdot)$ is the sigmoid function applied elementwise. That is, we are modelling the log-odds by a rank $k$ matrix approximation, where each dimension for $\alpha,\beta$ conceptually corresponds to a latent feature. As a result, we call this model the \emph{latent feature log-linear model} or \emph{LFL}. When $|\mathcal{Y}| > 2$, for the base class $y = R$, we are modelling the pairwise log-odds by a low-rank approximation:
$$\log \frac{p(y|x;w)}{p(R|x;w)} = \alpha^y_{r(x):} (\beta^y_{c(x):})^T.$$

Now suppose that we have side-information in the form of a vector $s(x)$ for a dyad $x$. Then, we augment the low-rank approximation with a linear discriminator, yielding
$$p(y | x; w) \propto \exp\left( \alpha^y_{r(x):} (\beta^y_{c(x):})^T + (w^y_s)^T s(x) \right).$$
Here, $w_s$ represents the weights used for side-information. Therefore, our model can use {one} \emph{or} {both} of latent features and side-information to make predictions.

The idea of extending multinomial logistic regression to incorporate information beyond the linear discriminator $w^T x$ is not entirely new, and has been studied in \emph{random effects} models in statistics e.g.\ \cite{RandomEffectsMLR}. Ours is the first contribution that applies a low-rank approximation of the log-odds for dyadic prediction, with the explicit aim of targetting both nominal and ordinal prediction tasks.


\subsection{ {Making predictions}}

Let $F(x;w)$ denote our model's prediction for the dyad $x$. A sensible choice for $F(x;w)$ will depend on the whether the outcomes in $\mathcal{Y}$ are nominal or ordinal.
	The mode, $\operatorname{argmax}_y p(y|x;w)$, is perhaps the only reasonable choice when the outcomes are nominal.
	The median, $F(x;w) = \text{median}(p(y|x;w))$ is sensible when the labels are ordinal, but has the disadvantage of being difficult to differentiate. If we further assume that the labels have a numeric structure, then the mean, $\mathbb{E}[y]=\sum_y yp(y|x;w)$, is a sensible alternative to the median. 
	

Just as the nature of the labels determines how we make predictions, it also determines how we train the model. We first present the objective function $f_\text{nom}$ minimized for nominal outcomes. Then we discuss how we can change this function to exploit the special structure present in ordinal outcomes, yielding a different function $f_\text{ord}$.

\subsection{ {Objective for nominal labels}}

When the $y$'s are nominal, a sensible objective function is the conditional log-likelihood (CLL) of the data. More generally, in place of CLL, we can use any \emph{proper loss} function, such as MSE \cite{ProperLoss}. Such functions are guaranteed to yield well-calibrated probability estimates. Since we are learning a large number of weights, it is beneficial to regularize our objective function with an $\ell_2$ penalty. Thus, the objective function we minimize is regularized negative CLL, which for a training set $\{(x_i,y_i)\}_{i=1}^n$ is
$$f_\text{nom}(w) = \frac{\lambda}{2}||w||^2 - \sum_{i=1}^n \log p(y_i|x_i; w).$$

\subsection{ {Objective for ordinal labels}}
\label{sec:train-ordinal}

For ordinal labels, one can train the model using $f_\text{nom}$, but this ignores the structure in the labels and reduces the model's predictive power. Instead, if we assume the labels have a numeric structure, it is more sensible to minimize the discrepancy between the true label and our model's prediction $F(x;w) = \mathbb{E}[y]$. A simple measure of discrepancy often used in collaborative filtering is the mean absolute error (MAE)
, defined as $\ell(y,\hat{y}) = |y - \hat{y}|$. For this choice, the regularized objective function is
$$f_\text{ord}(w) = \frac{\lambda}{2}||w||^2 + \sum_{i=1}^n |y_i - F(x_i;w)|.$$
Notice that MAE can be replaced with other loss functions, such as the MSE $\ell(a,b) = (a-b)^2$, or even the modified hinge loss of MMMF \cite{MMMF}; this does not change the underlying model, only the objective function.

We discuss a couple of points regarding the ordinal objective. First, if we measure test error using MAE it is sensible to optimize (regularized) MAE in training; however, this is not a proper loss function, which leads to uncalibrated probabilities. Second, if we use MAE as the training objective, ideally we would like to use $F(x;w) = \text{median}(p(y|x;w))$ (since the median is the solution to $\operatorname{argmin}_t \sum_i |x_i - t|$). As we noted earlier, however, this makes the objective non-differentiable. These issues represent tradeoffs that need to be carefully considered in practice.

There are other options when modelling ordinal data with a log-linear or logistic regression model, but these require modifying the underlying probability distribution \cite{OrdinalLR,Stereotype,OrdisticRegression}. Our focus is on using the same model as the nominal case, but modifying the training procedure to exploit the ordinal (or numerical) structure of labels.

\subsection{ {Strengths of the model}}

The LFL model meets all the goals for dyadic prediction listed in \S\ref{sec:dyadic-desiderata}. First, in contrast to most other dyadic prediction models, it can handle both nominal and ordinal $\mathcal{Y}$, just by changing the training objective. Second, it is easy for the model to handle side-information when it is present, which helps address the cold-start problem. Crucially, the model does {not} assume that \emph{only} side-information or only unique row/column object identifiers are relevant: it can make use of just one, or both of these pieces of information. Third, by virtue of it being a discriminative probabilistic model, it can produce well-calibrated probabilities and is resistant to sample-selection bias. Finally, the objective function is differentiable, and so we can use gradient based methods for training; in particular, we can use stochastic gradient descent (SGD), which scales to very large datasets.

\subsection{ {Weaknesses of the model}}

Unlike the basic log-linear model, the LFL objective function is not convex; it is only convex in $\alpha$ with $\beta$ fixed, and vice versa. Fortunately, our experimental results will demonstrate that it is easy to find good local optima.

Another observation is that we are learning a separate set of weights for each $y \in \mathcal{Y}$. If $|\mathcal{Y}|$ is large, then we will be learning many parameters, which increases the risk of overfitting. In nominal settings where there is truly no order to the various outcomes, then it is sensible that we have separate weights for each outcome. This is plausible even in the ordinal setting: e.g., the characteristics that make a user rate a movie $1$ star may be quite different from those that make her rate it $5$ stars. However, it is still desirable to exploit the ordinal structure to reduce the number of parameters. We briefly address this issue in \S\ref{sec:ordinal-lowrank}.


\section{ {Extensions to the model}}

We now discuss some important variations on our latent feature model. Specifically, we discuss adding bias weights, regularization, and reducing the number of parameters in the ordinal setting. We then show how one can apply the model for the special case of link prediction.

\subsection{ {Adding a bias weight}}
\label{sec:bias-weight}

In Equation \ref{eqn:latent-loglinear}, we did not have an explicit bias term $\gamma_y$. We can easily add a bias by forcing one latent feature in $\alpha$ and $\beta$ to have a constant value of $1$. 
This is equivalent to having a separate bias for each row and column object, and can be thought of as defining a scale for each of them. Notice that if $\alpha^y_{r:}$ has dimensionality $k$, two of its weights are related to bias terms (one weight is $1$, the other is the row bias). Henceforth, when we speak of a rank $k$ latent feature model, we mean one trained with $k+2$ parameters.

Our model can incorporate other tricks that have been used in standard matrix factorization models. For example, in the context of collaborative filtering, \cite{PMF} suggests imposing a prior on a user's latent weight vector that takes into account the identity of the movies she has rated. 

\subsection{ {Regularization}}

Earlier, we proposed regularization that penalized all components of $w$ equally. But it is plausible that the row and column weights have slightly different penalties. It is especially plausible that the weights for side-information be penalized differently. Also, as suggested in \cite{ImprovedMMMF}, it can be beneficial to apply the regularization inversely proportional to the (square root of the) number of observed entries for a particular row/column object i.e.\ the number of times that row object $r$ or column object $c$ appears as part of a dyad in the training set. This ensures that the penalty for row/column objects that have appeared only infrequently is larger than for those that have appeared frequently, which is sensible because we expect to overfit more for objects for which we have only limited data.


\subsection{ {Reducing parameters in the ordinal setting}}
\label{sec:ordinal-lowrank}

Our model keeps a separate set of weights for each $y \in \mathcal{Y}$, but it is plausible that these weights share some structure in the ordinal setting. One way to reduce the number of weights is to have a low-rank approximation of the latent weights themselves. This is inspired by the stereotype model for multinomial logistic regression \cite{Stereotype}: the difference here is that we apply it to the latent weights rather than the discriminator $w$. Essentially, we assume that for every label $y$, the weights can be decomposed as a linear combination of some set of base weights. The base weights are independent of $y$, but the scaling factors in the linear combination are not. Specifically, for any $y \in \mathcal{Y}$, we assume there are $\rho$ scalars $\phi^i$ such that
$$\alpha^y (\beta^y)^T = \phi^1_y \tilde{\alpha}^1 (\tilde{\beta^1})^T + \ldots + \phi^{\rho}_y \tilde{\alpha}^{\rho} (\tilde{\beta}^{\rho})^T.$$
Here, the weights $(\tilde{\alpha}^i, \tilde{\beta}^i)$ are shared among all outcomes $y$, and only the $\phi^i$'s vary. If $\rho \ll |\mathcal{Y}|$, this dramatically reduces the number of parameters to be learnt.

\subsection{ {Applying to link-prediction}}

As discussed earlier, in link prediction the row and column objects belong to the same space. Consider the setting where the input graph is unweighted and undirected, so that the matrix is symmetric and binary. To apply our latent feature model, we need to enforce symmetry: we need $p(y=1|(r,c);w) = p(y=1|(c,r);w)$. One way to achieve this is to maintain just one set of weights $\alpha$, for both the row and column objects.  Setting $y = 0$ to be the base class, we get the model
$$p(y = 1 | x;w) \propto \exp(\alpha_{r(x):} \alpha_{c(x):}^T).$$
As previously, define a matrix ${P}_{rc} = p(y=1|(r,c);w)$. Then,
${P} = \sigma(\alpha\alpha^T)$, 
where $\sigma(\cdot)$ is the sigmoid function, applied elementwise. That is, we take a symmetric low-rank approximation of the matrix $\sigma^{-1}(P)$.

One alternative is to use the decomposition $\alpha \Lambda \alpha^T$, where $\Lambda$ is a diagonal matrix. This uses the fact that any symmetric matrix has a decomposition $U\Lambda U^T$ for diagonal $\Lambda$; note however that this decomposition also requires $UU^T = I$, which is a difficult constraint to enforce when training our model. Our model is equivalent to learning ${P} = \sigma(\alpha \Lambda \alpha^T)$ when $\Lambda \succeq 0$, because we can absorb $\Lambda^{1/2}$ into $\alpha$. We found empirically that adding explicit scaling factors did not significantly improve the performance of our model.

For a directed graph, it is no longer appropriate to have a symmetric decomposition of ${P}$. Instead, we can return to our original model of having separate weights for row and column objects, which corresponds to weights for a users' incoming and outgoing edges. The drawback with this approach is that there is no sharing of information between the two sets of weights; they might as well belong to different users. One way to rectify this is to keep three sets of weights: one for the incoming edges, $\alpha$, one for the outgoing edges, $\beta$, and another for the global node information (i.e.\ weights that depend only on the identity of the node), $\gamma$. Then, we use the decomposition $\alpha \beta^T + \gamma \gamma^T$.

\subsection{Multi-relational data}

An important extension of the basic dyadic prediction task is where there are a series of outcomes. For example, in link prediction, there may be different types of links: ``is-family-member-of'', ``is-colleague-of'', et cetera. This is not the same as having nominal links, because every dyad can have many links of different types, rather than just one link with many possible outcomes. Therefore, it is akin to multi-label prediction rather than multi-class prediction. In this setting, we can extend our model to capture the structure underlying the various relations. Suppose we have $R$ binary relations, with outcomes $y^{1}, \ldots, y^{R}$. Then, for any $t \in \{1,\ldots,R\}$, we can model
$$p(y^{t} = 1 | (r, c)) = \sigma(\alpha_{r:} \Lambda^{t} \beta_{c:}^T ).$$
That is, we share the row/column object weights among all relations, but we give a different scaling factor depending on the relation. One can also imagine a model similar to the one we discussed in \S\ref{sec:ordinal-lowrank}, where
$$p(y^{t} = 1 | (r, c)) = \sigma\left( \sum_{q=1}^Q \lambda_q \alpha^q_{r:} (\beta^q_{c:})^T \right).$$
An interesting question is whether we can explicitly learn correlations among different relations, so that e.g.\ in a social network we can discover that ``is-colleague-of'' is positively correlated with ``is-friend-of''. This is an important question for future research.

\section{ {Experimental results}}

The experiments here demonstrate the flexibility of the LFL model proposed above, and its competitiveness with state-of-the-art methods on a range of different tasks. We present results targeting four important regimes: nominal data, ordinal collaborative filtering with and without side-information, and link prediction. In the process, we show our model scales to large datasets. We also present results on a matrix completion task involving handwritten digits.

In all experiments, we pick the regularization parameter(s) $\lambda$ using $3$-fold cross-validation on the training set. (We use different regularization for side-information and latent weights.)

For the medium-sized datasets, we train using LBFGS \cite{LBFGS}. Unlike first-order gradient methods, this does not require tuning of a learning rate, but at the price of higher computational cost. To demonstrate that the new model is scalable, results on large datasets are with stochastic gradient descent (SGD) as the training optimizer.

We report the mean and standard deviation of test set performance from $5$ runs of the training algorithm, where each run uses a different initialization of the weight vector. Different initializations lead to local optima of differing quality (recall that the optimization problem is nonconvex). Another potential source of variability is the split of training/test data: because each row/column object typically only has a few observations, some splits will be ``easier'' than others.

For the collaborative filtering tasks, we compared our method to MMMF, which is representative of most matrix factorization methods. We used the MATLAB code for this provided on the author's website\footnote{{\url{http://people.csail.mit.edu/jrennie/matlab/}}}. Note that this code does not include the suggestions in \cite{ImprovedMMMF}. The results reported for all other methods are the ones that appear in previously published experiments.

All experiments were run in MATLAB 2008b on a 2.67GHz Core i7 machine with 8 GB of RAM.

\subsection{ {Results on synthetic nominal data}}

We run experiments on a synthetic dataset to check that we can learn from nominal data, and that it is possible to find good local optima of the objective function despite its non-convexity. Another question is how well a method for ordinal labels like MMMF performs on this dataset. Intuitively, because such a method imposes an artificial structure on the outcomes, it will be difficult to learn a good model.

We constructed a matrix $M \in \mathbb{R}^{n \times n}$ whose entries were in $\{ 1, 2, 3 \}$. These can be thought of as indices into some set, e.g.\ \{bought, viewed, returned\}; the numeric encoding is just for convenience. We picked the entries of $M$ by sampling $M_{rc} \sim p(y|r,c) \propto \exp(\alpha^{y}_{r:} (\beta^y_{c:})^T)$, where there are $k = 5$ latent factors for the $\alpha$ and $\beta$ matrices. The entries of $\alpha, \beta$ were drawn uniformly at random from the interval $[-3, 3]$. We set some fraction of entries to be unobserved, which were used for testing, and let the remaining entries form the training set. The goal of training is to choose $\alpha,\beta$ that maximize the log-likelihood. Two parameters that will have an influence on the quality of the learned model are the size of the matrix, $n$, and the retention rate i.e.\ the fraction of entries kept for training.

The simplest measure of quality of our model is 0-1 accuracy. But this is only meaningful if we have a base error to measure against. Since we know the underlying probability distribution $p(y|x;w)$, we can find the Bayes error for a single matrix entry by $1 - \max_y p(y|r,c)$. If we are able to get close to the mean Bayes error, our model is doing well.

Results are presented in Table \ref{tbl:synthetic-results}. For varying choices of $n$ and the retention rate, our LFL model has high 0-1 accuracy. Our error rate is closest to the Bayes error when our training set is large: this is intuitive, because we expect the learning task to be simpler with more samples. Another promising result is that the accuracy of the LFL model increases with the size of the training data, \emph{despite} the increase in missing data. Also, as expected, MMMF does significantly worse than our model, demonstrating that an ordinal encoding is not sensible for this task.

\begin{table}[htb]
    \caption{0-1 Error of Rank $5$ Models on Synthetic Nominal Dataset.}
    \label{tbl:synthetic-results}	
	    \centering
	    \begin{tabular}{|ccc|cc|}
	        \hline
	        $n$ & Bayes & Retention & LFL & MMMF \\
	        \hline
	        & & $80\%$ & $7.7\%$ & $52.2\%$ \\ 
	        $500$ & $4.8\%$ & $50\%$ & $8.2\%$ & $53.0\%$ \\ 
	        & & $25\%$ & $12.0\%$ & $55.3\%$ \\ 
	        \hline
	        & & $80\%$ & $5.9\%$ & $48.9\%$ \\ 
	        $1000$ & $4.8\%$ & $50\%$ & $6.3\%$ & $53.2\%$ \\ 
	        & & $25\%$ & $8.1\%$ & $54.4\%$ \\ 
	        \hline
	        & & $80\%$ & $5.5\%$ & $47.2\%$ \\ 
	        $1500$ & $4.8\%$ & $50\%$ & $5.9\%$ & $47.3\%$ \\ 
	        & & $25\%$ & $7.0\%$ & $48.0\%$ \\ 
	        \hline
	    \end{tabular}
\end{table}
\subsection{ {Results on ordinal datasets}}
\label{sec:results-ordinal}

An important question is how our model compares with existing methods that target the ordinal setting. We focus on the canonical case of this problem, collaborative filtering. We emphasise that our goal is not to be the single best method for a collaborative filtering task, but rather to be competitive with existing methods while being more general than them.

\textbf{Results on \typewriter{bookcrossing}}. The {\small\typewriter{bookcrossing}} dataset
 consists of $1,149,780$ ratings made by $278,858$ users for $271,379$ books \cite{BookCrossing}. Following \cite{fLDA}, we pre-process the dataset to remove all ratings with the value $0$, users with less than $3$ ratings, and books with less than $6$ ratings. This left us with $342,464$ ratings over $35,689$ users and $138,660$ books. Unlike \cite{fLDA}, we did not use just the books with Amazon reviews; we also did not use any side-information, and just learned latent features.

Given the size of the dataset, we trained using SGD. We could not run the MMMF code, as it required too much memory. For our experiments, we performed three-fold cross-validation where each fold consists of $2/3$rds of each user's ratings. Our rank $5$ model attains an MAE of $1.0580 \pm 0.0028$, which is only slightly worse than the reported MAE for the fLDA method ($1.0317$) proposed in \cite{fLDA}. 
fLDA only uses side-information for making predictions: this information is hard to duplicate as it involves mining Amazon customer reviews. It is possible that our model can achieve higher accuracy when this extra side-information is incorporated as well. The training time of fLDA is not reported in \cite{fLDA}; our method processes the large dataset efficiently, and trains in around $8$ minutes.

\textbf{Results on $1M$ \typewriter{movielens}}. The 1M \typewriter{movielens} dataset consists of $1,000,209$ ratings for $6040$ users and $3900$ movies. With this dataset, we also address the scalability of our training algorithm by training with SGD. Following \cite{MMMF}, we randomly selected $5000$ users and for each user, picked a random rating and put it in the test set. The other ratings are put in the training set. We ended up with a training set of $836,865$ examples, each representing a user-movie rating dyad, and $5000$ test examples. Table \ref{tbl:scalability-results} presents the MAE and training time for our model and MMMF on the $1M$ dataset for various choices of latent factors $k$. We see that our model trains much faster than MMMF, while achieving competitive accuracy. In particular, our results for ranks $1$ and $5$ are statistically indistinguishable from MMMF. 

\begin{table}[htb]
    \centering
    \caption{Results on the $1M$ \typewriter{movielens} Dataset.}
    \label{tbl:scalability-results}    

    \begin{tabular}{|p{1.05in}|ccc|}
    \hline
    {Method} & $k$ & Test set MAE & {Train time} \\
    \hline
    MMMF & $1$ & $\textbf{0.6720} \pm \textbf{0.0000}$ & $35$ mins \\ 
    LFL & $1$ & $0.6777 \pm 0.0041$ & $7$ mins \\ 
    \hline
    MMMF & $2$ & $\textbf{0.6589} \pm \textbf{0.0009}$ & $40$ mins \\ 
    LFL & $2$ & $0.6635 \pm 0.0035$ & $7$ mins \\ 
    \hline
    MMMF & $5$ & ${0.6493} \pm {0.0001}$ & $72$ mins \\ 
    LFL & $5$ & $\textbf{0.6480} \pm \textbf{0.0059}$ & $14$ mins \\ 
    \hline
    \end{tabular}
\end{table}

\subsection{ {Results in the cold-start setting}}

Here, we present results showing that our model is able to learn to incorporate side-information easily, and make useful predictions in the cold-start setting. We took the 100K \typewriter{movielens} dataset, and randomly discarded $50$ users from the training set. These $50$ users are cold-start users when they appear in the test. In our experiments, we considered the following scenarios: (i) the standard setting, where there are no cold-start users or movies, (ii) there are cold-start users, but known movies, (iii) full cold-start, where both users and movies are unobserved. For (ii), we took the test set for (i) and then discarded all movies that occur in it from the training set. The side-information we used was the user's age, gender and occupation, and the movie's genre. 

We compare to a baseline method of predicting the mean from a latent feature model trained without side-information. That is, we train the standard latent feature model and for the dyad $(u,m)$ where $u$ was not present during training, our prediction is $\frac{1}{n_m} \sum_{(u',m) \in \mathcal{T}} F(u', m)$, where $n_m$ is the number of users in the training set $\mathcal{T}$ who have rated $m$, and $F(u',m)$ is the standard latent feature based prediction for dyad $(u',m)$. When both $u,m$ are not present in training, we just use the mean predicted rating for the entire training set.

When training with side-information, the following heuristic helps in avoiding bad local optima: first train the model without any side-information, and learn the latent weights. Then, use this as initialization to the model with side-information weights included i.e.\ initialize the latent weights for the second model to those learned by the first one. Better results are obtained when the latent weights were frozen for the second optimization i.e.\ we treat them as fixed, and just optimize the weights for the side-information. This is a form of block coordinate descent, where we optimize over two groups of variables by optimizing over each one in turn.

All models use rank $k=5$ and are optimized with LBFGS. Table \ref{tbl:side-info-results} summarizes our results. Learning with side-information significantly improves accuracy in the cold-start setting. Also, with side-information we do almost as well in the cold-start and standard settings; by contrast, the standard latent feature model does much worse in the cold-start setting. Comfortingly, side-information gives better MAE than the basic model when tested in the standard setting i.e.\ no cold-start users or movies. This means that taking side-information into account can give slightly better predictions than if we just learn latent features, which is as expected.

\begin{table}[htb]
    \centering
    \caption{Cold-start Results, Rank $5$ Model.}
    \label{tbl:side-info-results}
    
    \begin{tabular}{|p{0.935in}|p{1.05in}p{0.9in}|}
        \hline
        {Setting} & {Baseline MAE} & {Side-info MAE} \\
        \hline
        (i) Standard & $0.7162 \pm 0.0054$ & $\textbf{0.7063} \pm \textbf{0.0000}$ \\ %
        (ii) Cold-start & $0.8039 \pm 0.0000$ & $\textbf{0.7118} \pm \textbf{0.0208}$ \\ 
        (iii) Full cold-start & $0.9608 \pm 0.0000$ & $\textbf{0.7451} \pm \textbf{0.0196}$ \\
        \hline
    \end{tabular}
\end{table}

\subsection{ {Results on link prediction tasks}}

The \typewriter{coauthor} dataset \cite{coauthor} is a binary matrix indicating whether two authors have written a paper for NIPS together. The \typewriter{alyawarra} dataset \cite{alyawarra} contains kinship relations between people of the Alyawarra tribe in Australia. For both datasets, we use a random $80$-$20$ train-test set split: this means that for training, we assume that $20\%$ of entries in the matrix are missing, and we predict these at test time.

\textbf{Results on \typewriter{coauthor}}.
\label{sec:nips-coauthor}
Following \cite{IBP}, we focus on the $234 \times 234$ submatrix consisting of authors who collaborated with the most number of people. We report area under the ROC curve (AUC) as our performance measure, and compare our results to the ones given in \cite{IBP}, which studies the infinite latent factor model based on the Indian Buffet Process (IBP), the infinite relational model (IRM) and the mixed membership stochastic block model (MMSB). We also compare our method to MMMF, which is not explicitly designed for link prediction - it does not exploit the fact that the row and column spaces are the same - but helps us gauge how much improvement methods designed specifically for link prediction provide over standard collaborative filtering methods.

Our results are shown in Table \ref{tbl:link-prediction}. For all ranks, our model is superior to the MMSB and IRM models (whose ranks are not specified in the results of \cite{IBP}); however, we are slightly surpassed by the IBP. 
Also of interest in these results is that MMMF is outperformed by all other methods, which suggests that exploiting the special structure in link prediction tasks is essential to achieve good performance. MMMF gave slightly worse results for higher ranks, suggesting some overfitting. The random split of training/test set can have a significant impact on the accuracy of a learned model. We do not have the code or dataset partitions for the other methods, so this issue needs to be kept in mind when interpreting the results.

\begin{table}[htb]
    \centering
    \caption{AUC of Various Methods on \typewriter{coauthor} Dataset.}
    \label{tbl:link-prediction}
    
    \begin{tabular}{|c|cc|}
        \hline
        {Method} & $k$ & Test set AUC \\
        \hline
        MMSB & Unknown & $0.8705$ \\
        IRM & Unknown & $0.8906$ \\
        IBP & $20$ & $\textbf{0.9509}$ \\                
        \hline
        MMMF & $5$ & $0.8193 \pm 0.0132$ \\ 
        \hline   
        LFL & $5$ & $0.9235 \pm 0.0049$ \\ 
        LFL & $10$ & $0.9290 \pm 0.0165$ \\ 
        LFL & $20$ & $0.9424 \pm 0.0093$ \\ 
        \hline
    \end{tabular}    
\end{table}

\begin{figure}[htb]    
    \centering
    \includegraphics[scale=0.6]{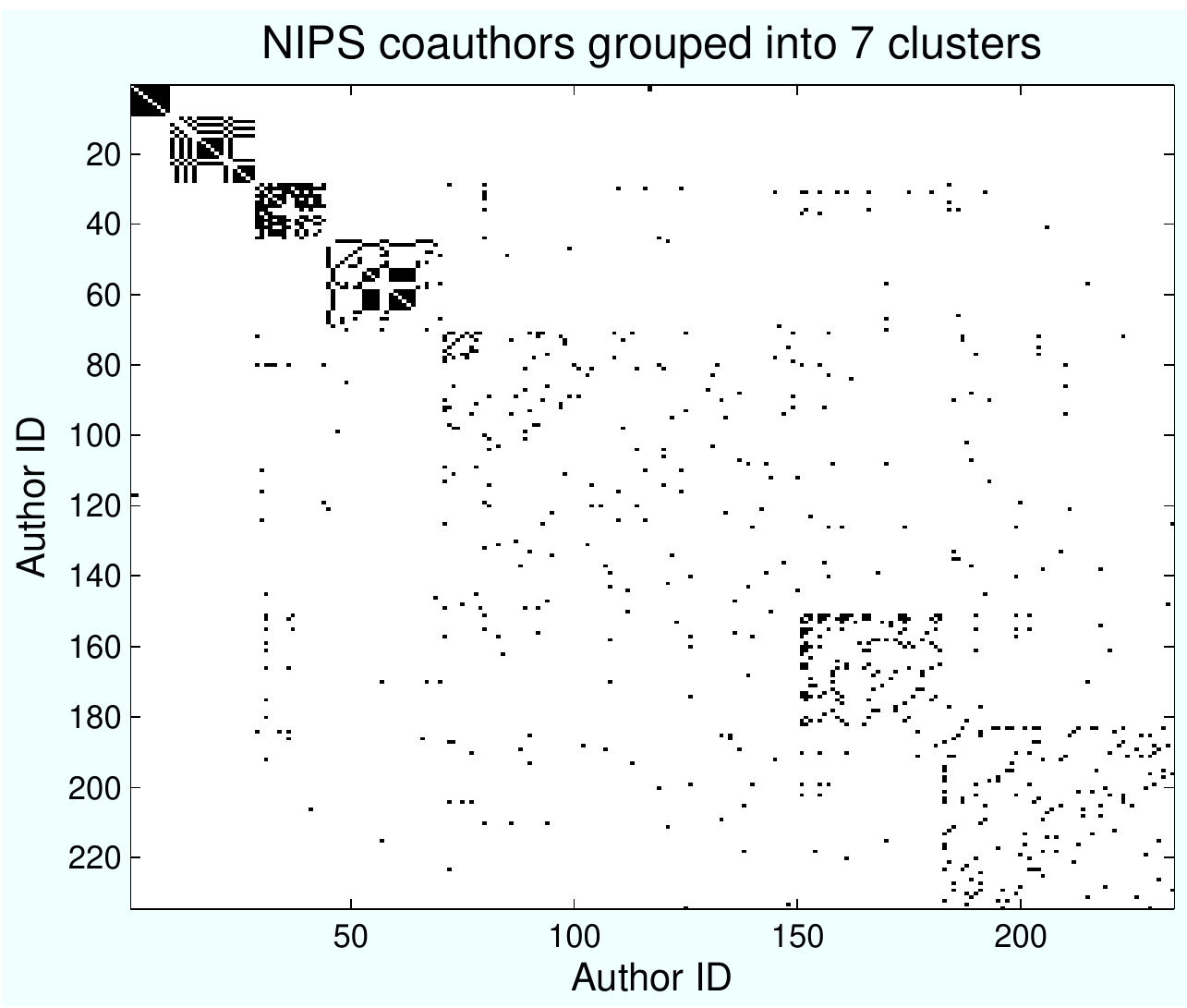}
		
		\
		
    \caption{\typewriter{coauthor} after clustering model output. Black/white entries indicate the presence/absence of a link.}
    \label{fig:nips-clusters}
\end{figure}

The learned weights for each author are useful for clustering. We applied $k$-means clustering on the user weights with $7$ clusters, and sorted the authors according to the learned clusters. Figure \ref{fig:nips-clusters} shows the resulting coauthor graph. We see that our user weights have identified significant cliques in the coauthor graph.

\textbf{Results on \typewriter{alyawarra}}. It is not sensible to run MMMF on this dataset, because the task is no longer binary, and so it is not sensible to impose an ordinal scale on the matrix entries. We can view this dataset as \emph{multi-relational}, so that each possible kinship relation defines a separate matrix, or as multi-category, where the number of classes is the number of kinship relations. We choose the latter because there is only one kinship relation observed between two people (that is, the outcomes are mutually exclusive). The reported results on this dataset \cite{IBP} use the multi-relational viewpoint. With this choice, each relation can have separate weights (a ``single'' model), or shared weights (a ``global'' model).

We ran our method for a number of ranks 
$k$, with log-likelihood as the objective function. The results in Table \ref{tbl:alyawarra} show that our rank $20$ model gives the best results of all methods. As we noted earlier, the precise train/test split of the data has a non-negligible impact on the final accuracy. Nonetheless, these results demonstrate that our model is certainly no worse than existing link prediction methods. 

\begin{table}[htb]
    \centering
    \caption{AUC of Various Methods on \typewriter{alyawarra} Dataset.}
    \label{tbl:alyawarra}
    
    \begin{tabular}{|c|cc|}
        \hline
        {Method} & $k$ & Test set AUC \\
        \hline
        MMSB global & Unknown & $0.8943$ \\
        IRM global & Unknown & $0.9143$ \\
        IBP global & $9$ & $0.9183$ \\
        \hline        
        MMSB single & Unknown & $0.9005$ \\        
        IRM single & Unknown & $0.9310$ \\
        IBP single & $9$ & ${0.9443}$ \\
        \hline
        LFL & $5$ & $0.9390 \pm 0.0006$ \\ 
        LFL & $10$ & ${0.9469} \pm {0.0013}$ \\ 
        LFL & $20$ & $\textbf{0.9475} \pm \textbf{0.0005}$ \\ 
        \hline
    \end{tabular}
\end{table}

\subsection{ {Results on matrix completion task}}

\begin{figure*}[htb]
    \centering
    \subfloat{\includegraphics[scale=0.3]{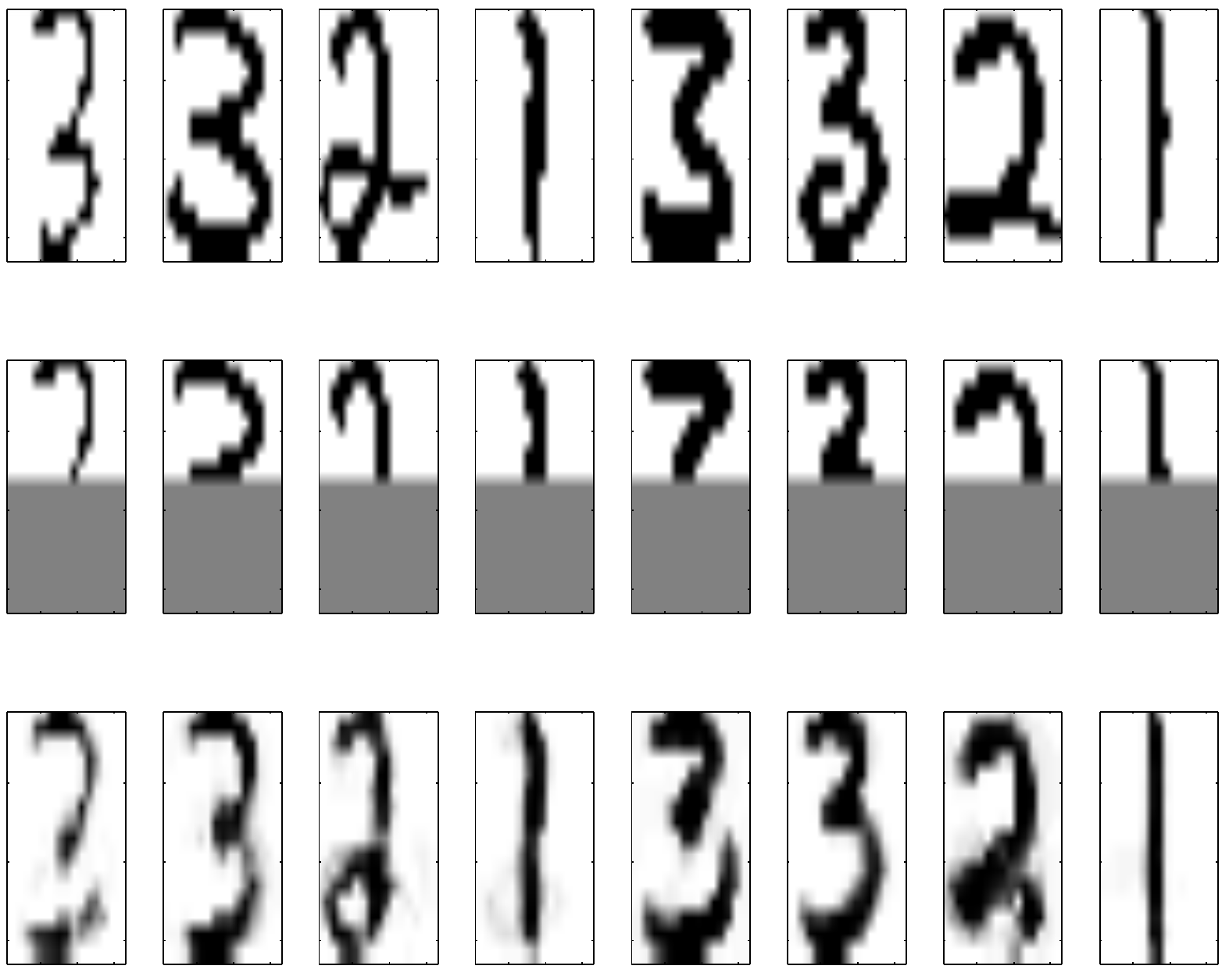}}
    \subfloat{\includegraphics[scale=0.3]{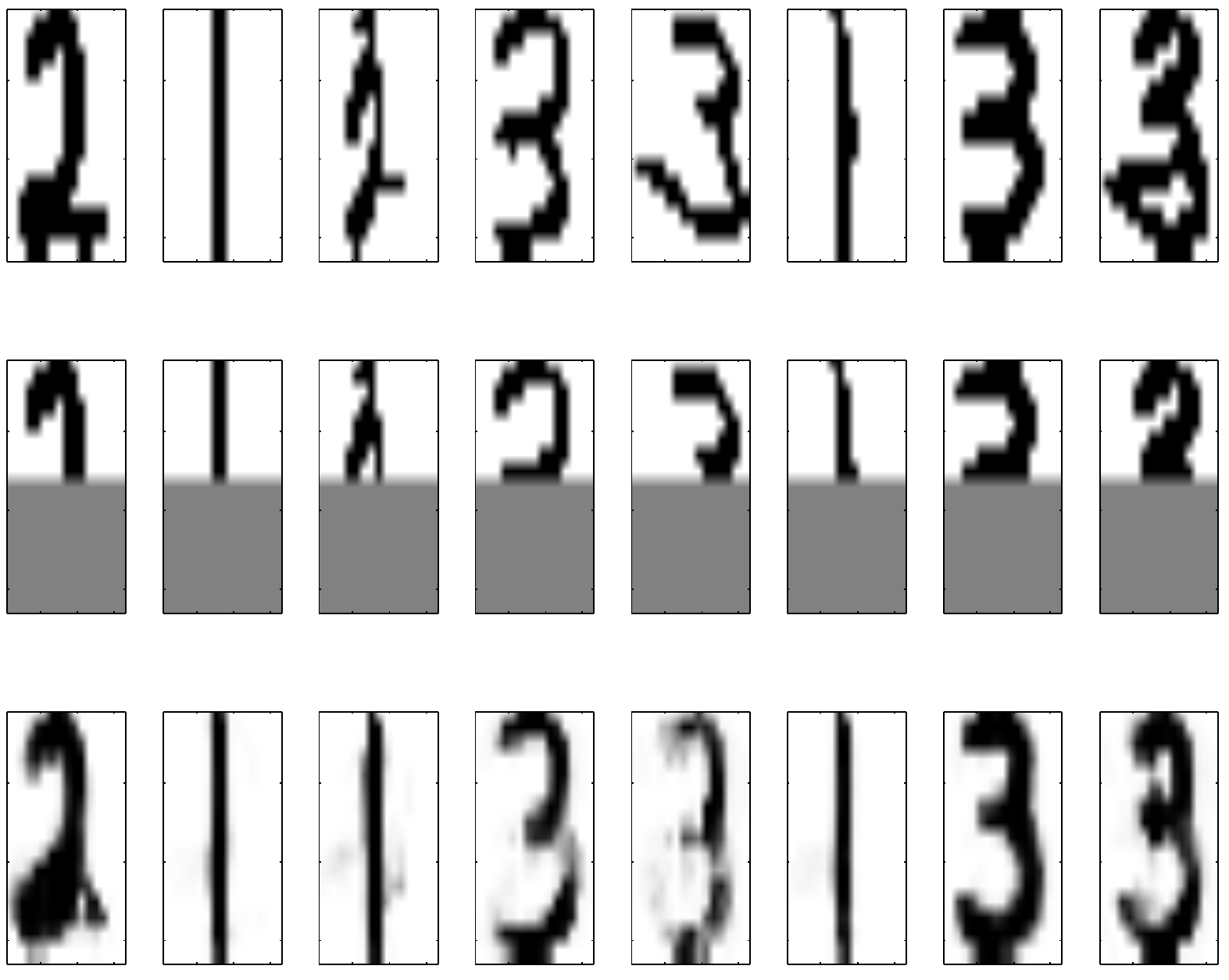}}   
    
    \
    
    \caption{\textbf{Top}: Original images; \textbf{Middle}: Corrupted images in training set; \textbf{Bottom}: Reconstruction using rank $3$ model.}
    \label{fig:digit-reconstruction}
\end{figure*}

We ran an experiment on the \typewriter{usps} dataset of handwritten digits, following \cite{InfiniteBinary}. We pick $100$ random images each of digits $1, 2$ or $3$. The images are binarized so that the pixel values are either $\pm 1$. For $16$ of these images, we chop off their bottom half by setting the pixel values to be ``missing''. The goal is to fill in the missing entries, or equivalently, to reconstruct the bottom half of the $16$ corrupted digits. This is a dyadic prediction problem where each digit is a row object, and each pixel position is a column object. We applied our LFL model on this dataset with rank $k=3$, optimizing MAE using LBFGS. Our results are shown in Figure \ref{fig:digit-reconstruction}: the top row shows the original versions of the $16$ corrupted images, the middle row shows the data as presented to the training algorithm, and the last row shows the predictions made by our model. (For this row, even the top half is the model prediction.) We see that it is able to accurately reconstruct most of the images. One exception is an image whose true digit is $3$, but which our model reconstructs as $1$; this behaviour is very understandable, because there is not enough information in the top half for even a human to predict the right answer.

\section{ {Conclusion}}

Existing methods for dyadic prediction are limited in one or more ways, most commonly in that they assume the labels lie on an ordinal scale, and only exploit either dyad members' unique identifiers or additional side-information. An ideal model should handle any of these settings. Other desirable properties for a dyadic prediction model are resistance to sample-selection bias and the generation of well-calibrated probabilities. Finally, given the size of datasets in real-world dyadic prediction tasks like collaborative filtering, it is essential for a model to be highly scalable. 
In this paper, we have presented a latent feature log-linear model (LFL) that addresses these issues. The model learns latent features even when we only have dyad identifiers, and can easily use side-information when it is available. We can apply it to very large datasets using stochastic gradient descent. It is a discriminative probabilistic method, and so has resistance to sample-selection bias and can generate well-calibrated probabilities. Our experiments show that we can learn from both nominal and ordinal labels, and that the model can use both dyad identifiers and side-information to achieve competitive accuracy to existing methods for collaborative filtering and link prediction. We also show that the model can be trained on large datasets using stochastic gradient descent.

\bibliographystyle{apalike}
\bibliography{references}

\end{document}